# GeoISF: Instance Semantic Forest Inspired Large-Scale Cross-View Geo-Localization via Ground LiDAR-to-Satellite Image

Di Hu, Xia Yuan*, *Member*, *IEEE*, and Chunxia Zhao

*Abstract*—**The problem of localization on a large-scale satellite image given a frame of query ground view point clouds remains challenging. Existing LiDAR-to-image cross-view localization methods struggle in large-scale scenarios due to limited semantic alignment and the modality gap between point clouds and satellite images. This paper introduces the large-scale LiDAR-to-image geo-localization pipeline called GeoISF. GeoISF introduces an instance semantic forest constructed using WordNet, which enhances temporal semantic representation and discriminative power by integrating semantic trees from multiple frames. By leveraging environmental semantic representation as a shared medium, GeoISF effectively bridges the modality gap and improves semantic matching accuracy. Extensive experiments demonstrate the superior performance of GeoISF in large-scale cross-view localization, 13.22 times better than the parallel LiDAR-to-image method in the R@10 metric on the KITTI dataset. The proposed method addresses the existing gap in large-scale LiDAR-to-image cross-view localization, offering a robust solution to the computational and accuracy challenges inherent in such scenarios. We will release the code as an open-source resource available online for the broader research community.**

## I. Introduction

Cross-view localization has attracted considerable research interest, which struggles with large-scale scenarios [1]. The satellite imagery employed in current LiDAR-to-image methods is constrained to a relatively narrow spatial extent, typically on the order of 250 meters by 250 meters [2]. In practical applications, pinpointing a specific location within a satellite image spanning several kilometers is a routine yet critical task, where the intricate nature of scenes presents significant challenges for cross-view localization. The dense semantic elements and repetitive road structures necessitate the development of efficient screening and retrieval mechanisms [3]. The core challenge in large-scale cross-view localization lies in effectively extracting and semantically matching environmental features. To bridge the gaps, we introduce the instance semantic forest (ISF) as a shared medium for ground-to-satellite semantic matching. In this paper, a novel ground-to-satellite geo-localization pipeline called GeoISF is proposed to locate the robot in a large-scale satellite image, marking the first attempt at LiDAR-to-image cross-view localization in large-scale scenarios.

Current large-scale cross-view localization algorithms primarily rely on matching query images with satellite image databases [4], achieving notable success in image-to-image matching. However, their ability to handle spatial structures and adaptability to lighting and weather conditions are not as good as those of LiDAR [5]. In contrast, LiDAR-to-image cross-view localization algorithms are relatively nascent, yet they have shown promising results, particularly in enhancing precision in complex scenarios [6]. Nevertheless, current LiDAR-to-image cross-view localization algorithms exhibit notable limitations due to the modality gap between point clouds and satellite images. Some approaches utilize local SLAM for pose estimation, requiring multi-frame image fusion and an initialization phase [2], [7]. Others primarily emphasize radar, highlighting potential enhancements in LiDAR-based fine-grained matching [8]. Moreover, several global localization algorithms, such as SaliencyI2PLoc [9], have been developed for aligning images with point clouds, which are limited to ground-to-ground scenarios.

Despite these efforts, no method for large-scale ground-to-satellite localization based on point clouds has been proposed yet, with the understanding and efficient processing of shared semantics remaining a significant limiting factor. Leveraging advancements in natural language generation models [10], text emerges as an alternative modality that can effectively extract key semantics, serving as an intermediary between images and point clouds to enable cross-modal feature matching [11]. In the proposed algorithm, text generation is applied to each cropped satellite image patch, aligning it with structured semantic labels. It allows for a certain degree of matching error during the progressive semantic distillation, eliminating patches that do not conform to road structures, thereby enhancing the accuracy of cross-view localization.

In the context of point cloud segmentation, perspective variations and scene occlusions significantly complicate the matching and retrieval processes [12]. However, the inherent containment relationships and spatial correspondences among instances can positively influence retrieval outcomes, with textual references providing valuable context for these spatial relationships. Motivated by these insights, this paper proposes the construction of the ISF based on WordNet [13], offering a semantically rich and comprehensive temporal semantic feature representation for point clouds. The instance semantic forest leverages the structural semantic ontology information extracted from point clouds, integrating semantic trees from multiple frames to enhance the temporal semantic representation. This approach not only improves the effectiveness of semantic information but also enhances its discriminative power, facilitating more accurate and robust

*(Corresponding author: Xia Yuan.)*

Di Hu, Xia Yuan, and Chunxia Zhao are with the School of Computer Science and Engineering, Nanjing University of Science and Technology, Nanjing 210094, China (e-mail: hudi@njust.edu.cn; yuanxia@njust.edu.cn; zhaochx@njust.edu.cn).

This research was funded by NingXia Academy of Agriculture and Forestry Sciences Science and Technology Innovation Guidance Technology Research Project, under grant NKYG-23-02.

cross-view localization. To summarize, the main contributions of this paper are threefold:

- A novel instance semantic forest-based localization pipeline called GeoISF is proposed in this paper. Leveraging shared semantics between ground and satellite, GeoISF effectively accomplishes large-scale cross-view geo-localization tasks by integrating point clouds and satellite imagery, marking the first attempt in this field.
- This paper introduces an innovative temporal semantic representation method that constructs a semantic forest through a tree-based ontology, integrating both geometric information and text description semantics. Compared to conventional approaches, the proposed method significantly enhances the capture of temporal dynamics and the effectiveness of semantic information based on semantic distillation.
- This paper introduces an innovative temporal semantic representation method that constructs a semantic forest through a tree-based ontology, integrating both geometric information and text description semantics. Compared to conventional approaches, the proposed method significantly enhances the capture of temporal dynamics and the effectiveness of semantic information based on semantic distillation.

## II. Related Works

### A. Large Scale Cross View Localization

Recently, cross-view localization has made significant progress. However, large-scale scenarios continue to receive little attention. LiDAR-to-image methods are pivotal in this domain [2], [7], but comprehensive studies on large-scale environments are still lacking. Dominating the field, imageto-image localization algorithms typically align ground-level images with satellite images through feature matching and similarity computation. Several methods utilize the particle filter to address the requirements of temporal localization, thereby enabling wide-area cross-view localization [14], [15]. However, the high precision of these algorithms necessitates prolonged robotic movement, often spanning kilometers. The other typical approaches are based on the goal of one-to-one image retrieval: given a ground image and a database of satellite images, determine which satellite image is the best match [16], [17]. These retrieval-based methods excel when training and testing sets are from the same region, but performance drops significantly in cross-area scenarios, highlighting the need for further advancements.

### B. Semantics-Based Scene Understanding

As a well-known lexical database, WordNet [13] has been utilized in different semantics-driven multimedia applications. WordNet is tree-structured, and each node of the tree comprises a set of words, each with one or more meanings. Each meaning has its synset, and three relationships (hypernyms, holonyms, and meronyms) are used to represent the relationship among a set of words. With the available semantic hierarchy for concepts and related semantic measures, WordNet has been applied in building benchmarks for various visual understanding tasks [18], which involve 2D object images, 3D models, and 2D scene images. As a prior related work, Yuan et al. proposed a semantic-tree-based approach for 3D scene model recognition [19]. The semantic tree is a hierarchical, directed graph, that utilizes semantic relatedness information between each scene object's label and the corresponding 3D scene's label. Furthermore, they also designed a novel method to construct a comprehensive scene semantic tree that integrates valuable scene semantic information, automatically encoding learned scene object occurrence probabilities within a scene category [13].

## III. Method

The framework of the proposed method is illustrated in Fig. 1. The algorithm hierarchically organizes nodes by leveraging textual labels derived from scene semantics within a single frame of point clouds, constructing an instance semantic tree (IST). Subsequently, by integrating ISTs across multiple timestamps, the ISF is established. Following the progressive semantic distillation process based on the ISF, the ground query point clouds are aligned with the filtered satellite imagery to determine the final position.

### A. Construction of Instance Semantic Forest

*1) Construction of IST:* Inspired by the hierarchical structure of humans looking for the location, we construct the instance-driven semantic forest $\mathcal{F}=\{T_1,T_2,...,T_k\}$ based on the spatial relationship. As illustrated in Fig. 1, the instance semantic tree $T_k$ in time $k$ is an instance-level hierarchical structure that organizes corresponding instances and their attributes based on the semantic hierarchy of WordNet synsets. The process commences by transforming the textual labels, extracted through panoptic segmentation based on Panoptic polarnet [20] and road structure extraction based on AVBM [12], into a richly attributed, dynamic graph representation. Both algorithms exhibit stable performance while satisfying real-time requirements. Each identified semantic instance (e.g., intersections, buildings) within the scene is instantiated as a node $n_i \in \mathcal{N}$. Each node $n_i$ is meticulously annotated with a descriptor tuple:

$$d_i = (c_i, f_i^{app}, f_i^{geom}, t, p_i)\,. \tag{1}$$

Here, $c_i$ denotes the semantic class grounded in the established ontology, $f_i^{app}$ encapsulates appearance-based features robust against viewpoint variations, and $f_i^{geom}$ captures critical intrinsic geometric properties. $t$ is the time stamp ensuring temporal coherency across sequential inputs, and $p_i$ represents the centroid position in the egocentric frame. The proposed method incorporates AVBM to directly identify the nearest unique road intersection ahead of the robot. This ensures node uniqueness in multi-intersection scenarios while excluding cases with no intersections. In GeoISF, the road intersection serves as a root node in the instance semantic tree, satisfying the criterion:

$$n_{root} = \{n_i \mid \phi_{topo}(f_i^{geom}) > \tau_{root}\}\,, \tag{2}$$

where $\phi_{topo}$ is a topological significance metric evaluating node centrality and structural connectivity [21], and $\tau_{root}$ is a learned threshold. Building upon this dynamically populated set of attributed nodes $\mathcal{N}$, the hierarchical extraction submodule orchestrates the formation of relational structures.

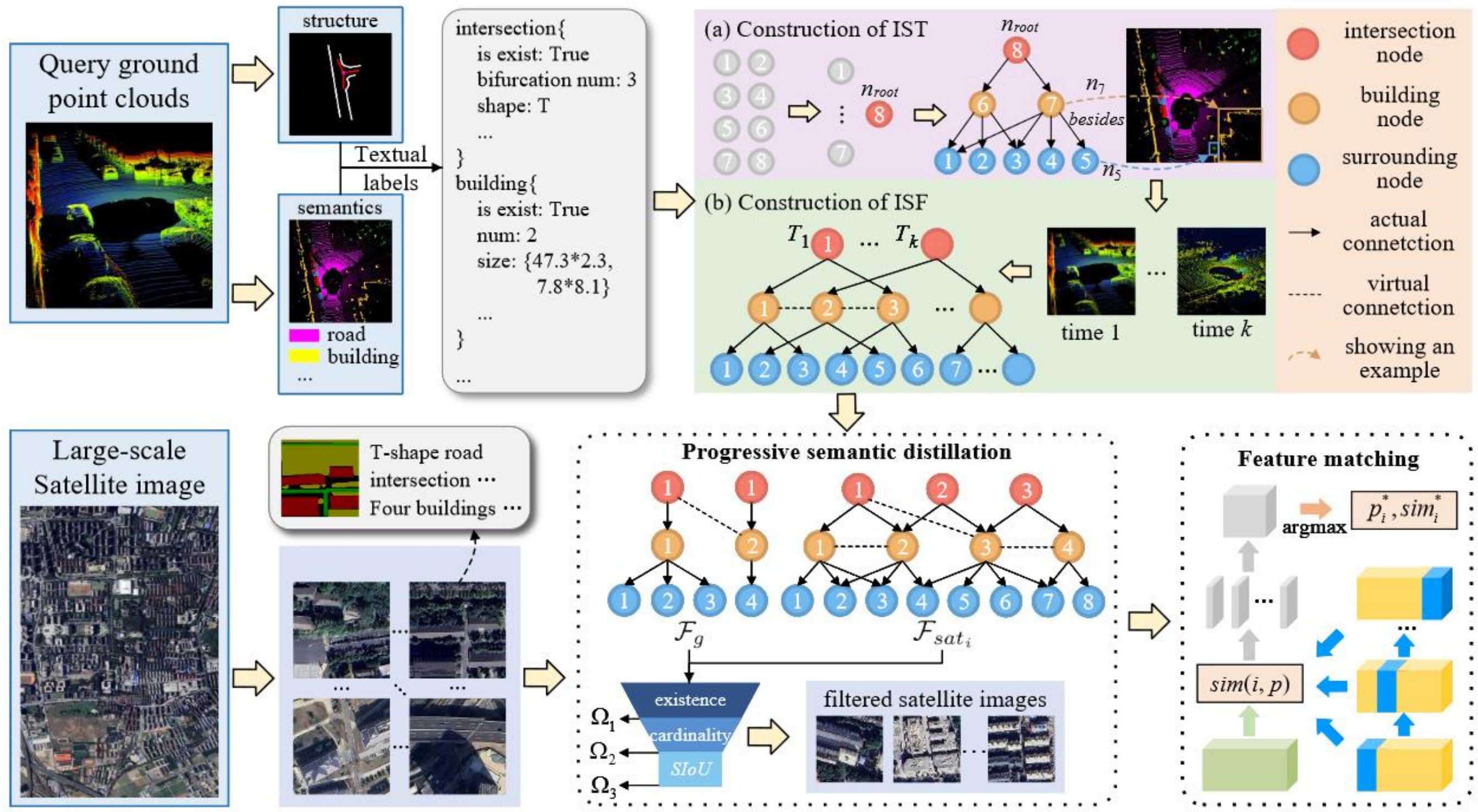

Fig. 1. Framework of the proposed method. The algorithm first extracts textual labels through panoramic segmentation and road structure detection, and constructs a temporal semantic forest based on semantic ontology. For large-scale satellite image, the initial step involves uniformly cropping the images both horizontally and vertically. The cropped patches cover an approximate spatial extent of 200 meters by 200 meters. Subsequently, SRSS-based semantic segmentation and ChatGPT-based text generation are applied on each satellite patch. On this basis, a progressive semantic distillation method is designed to reduce the number of satellite images to be matched, and feature matching is performed with point cloud semantics in the filtering results, taking the group with the highest similarity as the final result.

It establishes semantic edges $e_{ij}^{act}$ between nodes $n_i$ and $n_j$ based on contextual predicates reflecting real-world adjacency, containment, and functional association.

$$s_{ij}^{act} = \varphi([f_i \oplus f_j], \Delta p_{ij}, \Delta t),$$
$$e_{ij} = \begin{cases} 1, & if\ s_{ij}^{act} > \tau_{edge} \\ 0, & otherwise \end{cases}. \quad (3)$$

Here, $f_i$ and $f_j$ denote the geometric features of nodes $n_i$ and $n_j$, while $s_{ij}^{act}$ represents the similarity between them, determining whether they are connected in the IST. $\Delta p_{ij}$ and $\Delta t$ refer to the spatial distance and temporal difference between the two nodes, respectively. In addition, $\varphi : \mathbb{R}^k \rightarrow [0,1]$ is a similarity metric modeled as:

$$\varphi(\cdot) = \sigma(\mathbf{W}_2 \operatorname{ReLU}(\mathbf{W}_1[f_i \oplus f_j \oplus \| p_i - p_j \|_2 \oplus | t_i - t_j |])), \quad (4)$$

where $\sigma$ is the sigmoid function, $\mathbf{W}_1$ and $\mathbf{W}_2$ are learnable weights, $\oplus$ denotes concatenation, and $\tau_{edge}$ is a sparsity inducing threshold. By incorporating geometric features and positional discrepancies between two nodes, the computed similarity effectively assesses their spatial coincidence and geometric relationships. Nodes spatially proximate or functionally linked form directed edges $e_{ij}$ from parent $n_i$ to child $n_j$. Starting from root nodes, trees grow iteratively by connecting child nodes to their most probable parent:

$$T_k = \bigcup_{\forall n_j \notin T_k} \arg\max s_{ij}^{act}, \quad (5)$$

where $k$ denotes various timestamps. In this strategy, the intersection nodes typically serve as the root nodes of the hierarchical structure, with building nodes occupying the second level and surrounding nodes designated as leaf nodes, thereby forming a tree structure as depicted in Fig. 1.

*2) Construction of ISF:* Based on IST, the dynamic construction incorporates the concept of virtual connections to implicitly model potential or non-explicit spatial-semantic relationships that cannot be directly ascertained from immediate adjacency or containment rules. Each IST corresponds to a frame of point clouds. These connections, denoted $e_{ij}^{virt}$, are probabilistically established between node pairs $(n_i, n_j)$ based on the compatibility and similarity of their respective descriptor tuples $d_i$ and $d_j$, utilizing a learned similarity metric $\phi(d_i, d_j)$. Specifically, the probability is defined as:

$$P(e_{ij}^{virt} = 1) = \sigma_e(\phi(d_i, d_j; \Theta)), \quad (6)$$

where $\sigma_e$ is a sigmoid function, and $\Theta$ parameterizes the model evaluating feature coherence. These virtual edges $e_{ij}^{virt}$ are not enforced physical links but represent hypothesized topological continuities within the scene structure. This mechanism provides an essential capacity to represent fragmented or partially occluded entities and anticipate latent spatial-semantic relationships critical for holistic scene interpretation when explicit hierarchical structuring occurs.

The instance semantic forest at various timestamps is constructed as $\mathcal{F}=\{T_1,T_2,...,T_k\}$. This hierarchical aggregation intrinsically encodes scene semantics through conditional dependencies, where higher-level nodes govern contextual relationships of subordinated elements.

### *B. Semantics-Driven Cross View Image Retrieval*

*1) Progressive Semantic Distillation:* To navigate vast geospatial data efficiently, the approach employs progressive semantic distillation, leveraging the inherent hierarchies within the instance semantic forest $\mathcal{F}_g$. Analogous to the construction of instance semantic forest from ground multi-frame point clouds, instance semantic forests are similarly generated from satellite image patches with semantic extraction. ChatGPT 4.0 is leveraged to produce descriptive textual representations of satellite imagery, where specific generation templates are established to regulate the output structure. The descriptor tuple is denoted as $d_{si}=(c_{si},f_{si},p_{si})$, where $c_{si}$ represents the semantic class, $f_{si}$ means the extracted features, and $p_{si}$ denotes the position. The connection is defined as $s_{ij}^{sat}=\varphi(f_i,f_j,\Delta p_{ij})$ based on the spatial relationship. Through the combination of nodes and hierarchical connections, the instance semantic forest $\mathcal{F}_{sat}$ is constructed for the semantic distillation.

As an alternative to computationally intensive brute-force matching, it proposes a strategic cascade of abstractions. Given that both satellite image text generation and point cloud semantic extraction rely on real-world physical measurements, the derived $\mathcal{F}_g$ and $\mathcal{F}_{sat}$ are comparable without the need for scale alignment. Given the point clouds $P$ and the satellite image database $S_{raw}=S_1,S_2,...,S_N$, the constructed ISFs are denoted as $\mathcal{F}_g$ and $\mathcal{F}_{sat_i}(i=1,2,...,N)$. The distillation framework progressively reduces the candidate space through hierarchical constraints, leveraging structural isomorphism between $\mathcal{F}_g$ and $\mathcal{F}_{sat_i}$. Initial filtering enforces node existence alignment, requiring:

$$\forall c_k,\exists d_{si}\in\mathcal{F}_{sat_i}:c_{si}>c_k, \tag{7}$$

yielding $\Omega_1\subseteq\Omega$. Here, $c_{si}$ represents the semantic class in $\mathcal{F}_{sat_i}$, while $c_k$ denotes the semantic class in $\mathcal{F}_g$. $c_{si}>c_k$ indicates that the semantic category in $\mathcal{F}_{sat_i}$ encompasses that in $\mathcal{F}_g$. Subsequent cardinality matching imposes:

$$\forall c_k,|\{d_g\in\mathcal{F}_g\mid c_g=c_k\}|<|\{d_{si}\in\mathcal{F}_{sat_i}\mid c_{si}=c_k\}|, \tag{8}$$

generating $\Omega_2\subseteq\Omega_1$. The semantic class in the descriptor tuple $d_g$ is represented by $c_g$. In addition, a critical prerequisite is the semantic coherence constraint, which ensures spatial compatibility. Therefore, we introduce the Semantic Intersection-over-Union (SIoU):

$$SIoU(\mathcal{F}_g,\mathcal{F}_{sat_i})=\frac{|\mathcal{F}_g\cap\mathcal{F}_{sat_i}|}{|\mathcal{F}_g\cup\mathcal{F}_{sat_i}|}\cdot\rho(\Delta\mathcal{F}_g,\Delta\mathcal{F}_{sat_i}), \tag{9}$$

where $\rho$ computes structural consistency via edge gradient correlation [22]. This discards geometrically incompatible candidates, and generates $\Omega_3\subseteq\Omega_2$. The process of semantic distillation progressively prunes the candidate space while preserving localization fidelity, collapsing the expansive satellite database $S_{raw}$ into a sparse set of semantically coherent anchor points $S_{sparse}$. Final localization then resolves efficiently within this distilled, high-likelihood set.

*2) Feature Matching:* Following the semantic distillation, the fine-grained matching establishes geometric correspondence between the point clouds $P$ and the filtered satellite image database $S_{sparse}=S_1,S_2,...,S_S$. Given the extracted feature vector $f_P$ for the point clouds and $f_{si}$ for the satellite images, we formulate matching as a maximization of normalized feature correlation to quantify alignment precision. For $S_i$, a sliding-window search is executed over $P$ within a bounded displacement window $W$ centered on prior filtering outputs. At each candidate position $p=(x,y)\in W$, a local patch $R_P(p)$ is extracted from $P$. The similarity between $S_i$ and $R_P(p)$ is evaluated via the cosine similarity of their flattened feature vectors:

$$sim(i,p)=\frac{vec(f_{si})\cdot vec(f_{R_P(p)})}{\|vec(f_{si})\|_2\cdot\|vec(f_{R_P(p)})\|_2}, \tag{10}$$

where $vec(\cdot)$ denotes vectorization. $sim(i,p)$ measures angular consistency in high-dimensional semantic space, rendering the comparison robust to nonlinear intensity variations between ground and aerial perspectives. The optimal alignment position $p_i^*$ and its score $sim_i^*$ for $S_i$ are derived as:

$$p_i^*,sim_i^*=\arg\max_{p\in W} sim(i,p). \tag{11}$$

The global matching solution is then determined by ranking the similarity scores $sim_1^*,...,sim_N^*$. The satellite image $S_k$ and position $p_k^*$ associated with the top-ranked score constitute the final matched pair, achieving high-precision geographic coordinates through the semantic feature space correlation paradigm. This strategy effectively bridges cross-modal domain gaps while preserving spatial discriminability.

## IV. Experiments

### *A. Experimental Data and Setup*

The experiment was conducted using two distinct datasets: KITTI-raw [23] and NCP-Intersection [12]. The point clouds in the KITTI dataset were collected via the Velodyne HDL64E LiDAR, whereas those in the NCP-Intersection were obtained using a Velodyne VLP-16 LiDAR. Given the absence of a dedicated dataset for large-scale LiDAR-to-image cross-view localization, we have refined the satellite image coverage using two existing datasets. In contrast to current cross-view localization datasets, which are limited to satellite imagery of a few intersections, our newly acquired largescale satellite images encompass hundreds of intersections.

We follow Congeo [16] and FRGeo [24] in using the recall accuracy at top K and top 1% as the evaluation metric for our method. A retrieval is deemed correct if the corresponding satellite image is included within the retrieved set. Given that the metric is limited to retrieval accuracy, discrepancies in orientation angle are excluded from consideration. To assess the performance of the proposed method, we performed extensive comparative analysis against 13 state-of-the-art approaches on the KITTI and NCP-Intersection datasets. These comparative methods encompass 8 image-to-image techniques and 5 LiDAR-to-image methodologies. The former category comprises established cross-view localization algorithms rooted in image retrieval, including SAFA [25], VIGOR [26], TransGeo [27], SAIG [28], GeoDTR [29], GeoDTR+ [30], FRGeo [24], and Congeo [16]. The latter category includes Zhang et al. [31], Sun et al. [32], Wang et al. [33], Hu et al. [2], and SCLM [7]. While the primary focus of Zhang et al. [31], Sun et al. [32], and Wang et al. [33] was initially curb and intersection detection, these methods have been subsequently enhanced in Hu et al. [2] to facilitate cross-view localization tasks. It should be noted that the majority of the comparative algorithms are specifically designed for image retrieval-based cross-view localization and are not inherently tailored to address localization tasks in large-scale scenes. Nevertheless, the retrieved satellite image index can be effectively utilized as a foundational reference for localization within extensive satellite imagery datasets.

TABLE I
COMPARISONS BETWEEN THE PROPOSED AND STATE-OF-THE-ART METHODS ON THE KITTI ANDD NCP-INTERSECTION DATASETS. BEST AND SECOND BEST RESULTS SHOWN IN BOLD AND UNDERLINE, RESPECTIVELY.

| Type | Method | KITTI (64-channel) | | | | NCP-Intersection (16-channel) | | | |
|---|---|---|---|---|---|---|---|---|---|
| | | r@1 | r@5 | r@10 | r@1% | r@1 | r@5 | r@10 | r@1% |
| Image-to-image | SAFA [25] | 1.15 | 7.97 | 14.77 | 40.52 | 0.58 | 5.42 | 12.63 | 35.85 |
| | VIGOR [26] | 8.31 | 19.24 | 28.64 | 55.47 | 5.27 | 15.42 | 25.05 | 52.93 |
| | TransGeo [27] | 14.92 | 30.41 | 42.83 | 69.25 | 11.77 | 24.38 | 34.95 | 57.62 |
| | SAIG [28] | 14.87 | 32.34 | 47.82 | 72.68 | 14.27 | 28.09 | 37.52 | 63.82 |
| | GeoDTR [29] | 20.42 | 41.89 | 53.71 | 77.37 | 18.38 | 39.92 | 49.25 | 74.17 |
| | GeoDTR+ [30] | 22.08 | 44.62 | 59.75 | 79.20 | 21.72 | <u>43.78</u> | <u>56.01</u> | 76.03 |
| | FRGeo [24] | 23.16 | 40.93 | 54.14 | 71.02 | <u>22.11</u> | 39.83 | 52.80 | 70.19 |
| | Congeo [16] | <u>27.74</u> | <u>46.33</u> | <u>62.81</u> | <u>80.64</u> | **27.69** | **48.20** | **59.76** | **81.95** |
| Image-to-LiDAR | Zhang et al. [31]+Hu et al. [2] | - | - | 2.37 | 15.92 | - | - | 1.15 | 11.74 |
| | Sun et al. [32]+Hu et al. [2] | - | - | 2.88 | 18.63 | - | - | 1.64 | 12.81 |
| | Wang et al. [33]+Hu et al. [2] | - | - | 3.06 | 19.62 | - | - | 2.31 | 14.40 |
| | Hu et al. [2] | - | - | 5.48 | 30.87 | - | - | 3.96 | 18.74 |
| | SCLM [7] | - | - | 2.19 | 17.36 | - | - | 1.58 | 13.02 |
| | Proposed | **31.35** | **52.92** | **72.44** | **91.53** | 20.65 | 39.81 | 54.73 | <u>77.09</u> |

TABLE II
ABLATION STUDY ON VARYING NUMBERS OF LIDAR SCAN CHANNELS IN KITTI DATASET. (BOLD: BEST)

| Method | 64-channel | | | | 32-channel | | | | 16-channel | | | |
|---|---|---|---|---|---|---|---|---|---|---|---|---|
| | r@1 | r@5 | r@10 | r@1% | r@1 | r@5 | r@10 | r@1% | r@1 | r@5 | r@10 | r@1% |
| Zhang et al. [31]+Hu et al. [2] | - | - | 2.37 | 15.92 | - | - | 1.87 | 14.14 | - | - | 1.28 | 11.36 |
| Sun et al. [32]+Hu et al. [2] | - | - | 2.88 | 18.63 | - | - | 2.29 | 15.84 | - | - | 1.81 | 12.92 |
| Wang et al. [33]+Hu et al. [2] | - | - | 3.06 | 19.62 | - | - | 2.71 | 16.22 | - | - | 2.49 | 15.21 |
| Hu et al. [2] | - | - | 5.48 | 30.87 | - | - | 4.70 | 23.13 | - | - | 4.39 | 20.13 |
| SCLM [7] | - | - | 2.19 | 17.36 | - | - | 1.95 | 15.18 | - | - | 1.73 | 13.76 |
| Proposed | **31.35** | **52.92** | **72.44** | **91.53** | **24.95** | **44.19** | **68.06** | **85.64** | **18.52** | **40.68** | **52.08** | **75.19** |

TABLE III
THE FULL TABLE OF ABLATION STUDIES ON FOV=90°, 180°, AND 360° OF LIDAR IN THE KITTI DATASET. (BOLD: BEST)

| Method | FoV=360° | | | | FoV=180° | | | | FoV=90° | | | |
|---|---|---|---|---|---|---|---|---|---|---|---|---|
| | r@1 | r@5 | r@10 | r@1% | r@1 | r@5 | r@10 | r@1% | r@1 | r@5 | r@10 | r@1% |
| Zhang et al. [31]+Hu et al. [2] | - | - | 2.37 | 15.92 | - | - | 0.57 | 5.91 | - | - | - | 0.95 |
| Sun et al. [32]+Hu et al. [2] | - | - | 2.88 | 18.63 | - | - | 1.25 | 7.23 | - | - | - | 1.12 |
| Wang et al. [33]+Hu et al. [2] | - | - | 3.06 | 19.62 | - | - | 1.43 | 9.06 | - | - | - | 0.77 |
| Hu et al. [2] | - | - | 5.48 | 30.87 | - | - | 3.62 | 18.51 | - | - | 1.12 | 5.89 |
| SCLM [7] | - | - | 2.19 | 17.36 | - | - | 1.82 | 14.85 | - | - | 0.86 | 6.01 |
| Proposed | **31.35** | **52.92** | **72.44** | **91.53** | **16.74** | **39.16** | **50.12** | **68.44** | **2.29** | **7.32** | **15.33** | **31.08** |

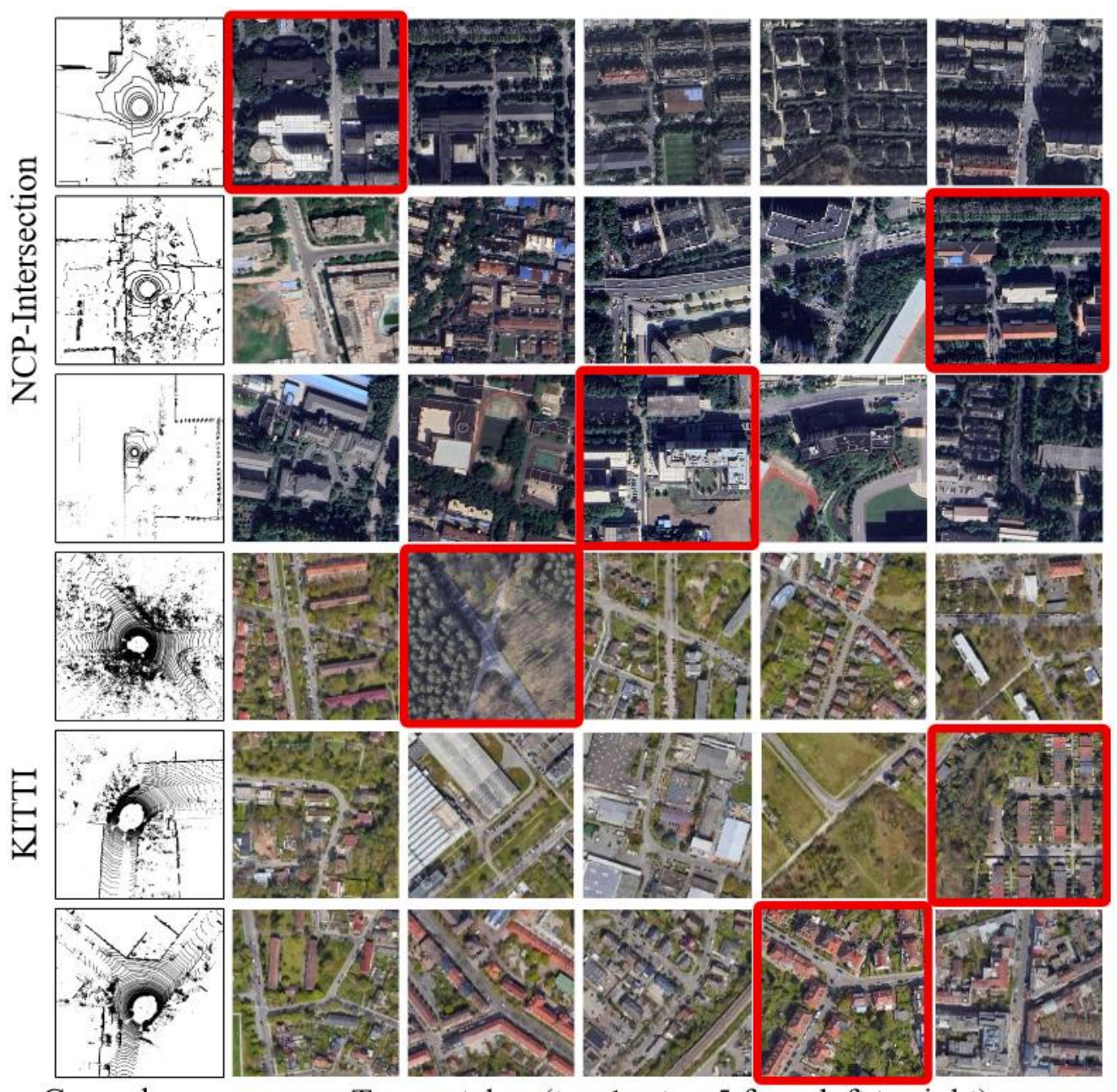

Fig. 2. Image retrieval examples on NCP-Intersection and KITTI dataset. The satellite image bordered by red square is the groundtruth.

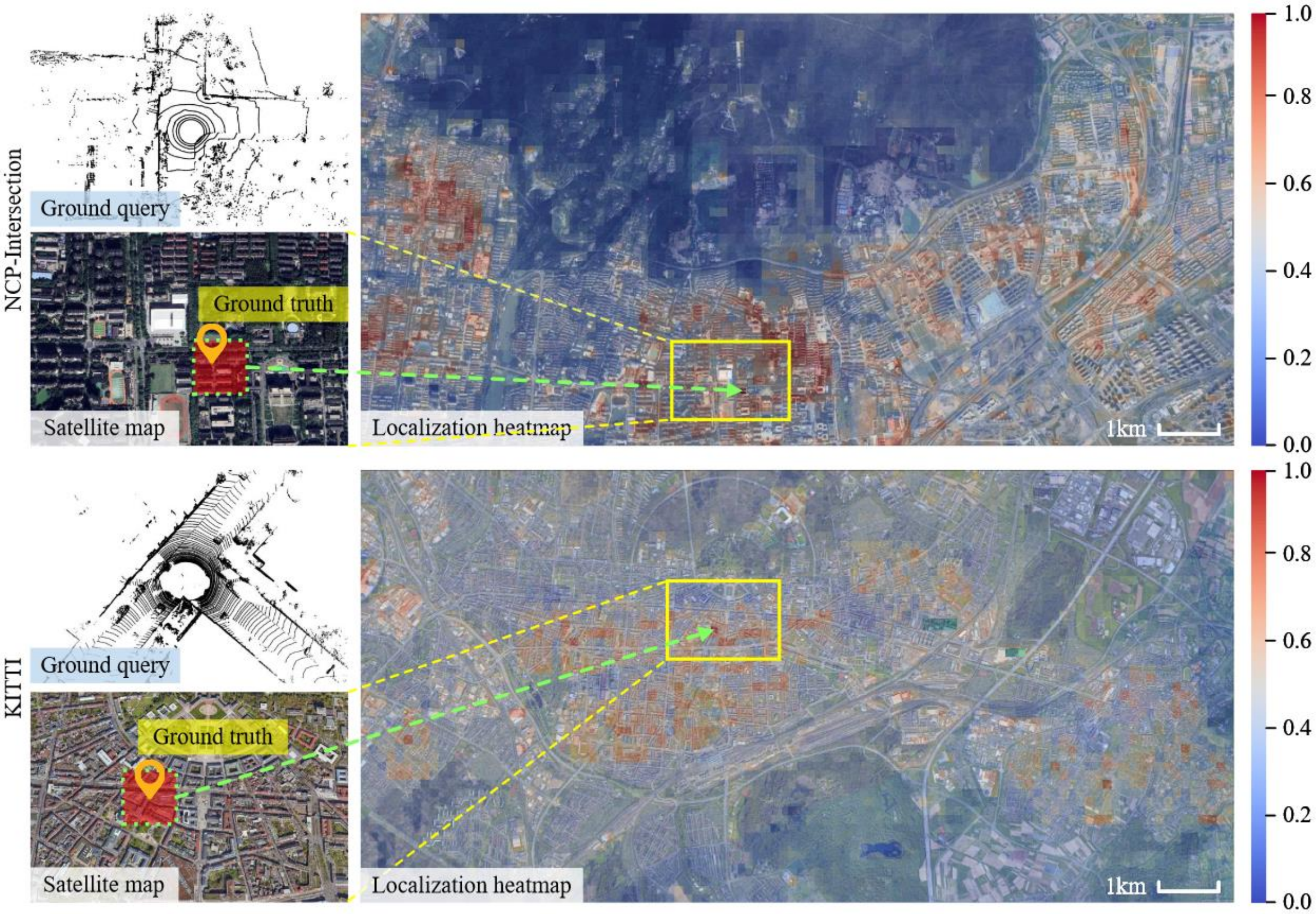

Fig. 3. Large-scale geo-localization examples on KITTI and NCP-Intersection datasets.

### *B. Evaluation Results*

To validate the effectiveness of the proposed algorithm, we performed comprehensive comparative experiments utilizing the KITTI and NCP-Intersection datasets, as detailed in Table I. The experiments are rigorously designed to operate under cross-area conditions. The proposed method emphasizes the spatial attributes of ground query point clouds, demonstrating superior efficacy across all settings. As presented in Table I, the optimal comparison algorithm for the image-to-image approach achieves an r@1% value of 80.64% on the KITTI dataset, whereas its counterpart in the image-to-radar paradigm attains a significantly lower value of 30.87%. The primary reason lies in the fact that existing LiDAR-to-image cross-view localization algorithms predominantly concentrate on 3-DoF pose estimation, with minimal emphasis on retrieving query images from databases. This fine-grained pose matching paradigm is inherently ill-suited for the task of geo-localization in large-scale scenarios.

In contrast, the proposed method relieves the critical gap in large-scale LiDAR-to-image cross-view localization. As demonstrated in the performance of the KITTI dataset, our approach significantly outperforms comparable algorithms. In comparison to the reproduced LiDAR-to-image algorithms, the proposed algorithm demonstrates significant enhancements, achieving a 2.97-fold improvement in the r@1% metric and a 13.22-fold improvement in the R@10 metric. Furthermore, it exhibits substantial gains in the R@1 and R@5 metrics, with enhancements of 31.35% and 52.92% from zero, respectively. Furthermore, when benchmarked against state-of-the-art image-to-image algorithms, our method demonstrates a notable improvement of 10.89% in r@1% and 9.63% in r@10 under cross-area conditions.

On the NCP-Intersection dataset, the proposed method maintains a substantial performance advantage over the LiDAR-to-image approaches, achieving a 4.11-fold enhancement in the r@1% metric and a 13.82-fold improvement in the R@10 metric. Additionally, it achieves notable gains in the R@1 and R@5 metrics, with improvements of 20.65% and 39.81% from zero, respectively. Compared to image-to-image algorithms, the proposed method demonstrates superior performance over most existing approaches, but underperforms compared to Congeo [16], which achieves a r@1% of 81.95%. In contrast to the KITTI dataset, NCP-Intersection features sparser point clouds. Specifically, the point cloud in KITTI comprises 64 channels, whereas the NCP-Intersection dataset contains only 16 channels. This disparity leads us to hypothesize that LiDAR density significantly influences the algorithm's efficacy, explaining the observed underperformance of the proposed algorithm compared to Congeo on NCP-Intersection dataset.

To validate this hypothesis, we utilized the 64-channel data from the KITTI dataset as a baseline and conducted ablation experiments by extracting point clouds with 32 and 16 channels, with the corresponding results presented in Table II. The experimental results demonstrate a significant correlation between point cloud density and algorithm performance within the same dataset. Specifically, as the point clouds become sparser, the algorithm's performance deteriorates. The r@1% metric for the 16-channel LiDAR is 75.19%, while the r@1 metric is 18.52%. This phenomenon can be attributed to the impact of sparse point clouds on semantic extraction, which subsequently leads to inaccuracies in constructing the semantic forest. Notably, when utilizing a 32-channel LiDAR configuration, the r@1% improves to 85.64% and the r@1 increases to 24.95%, achieving performance comparable to that of Congeo, as evidenced by the data in Table I.

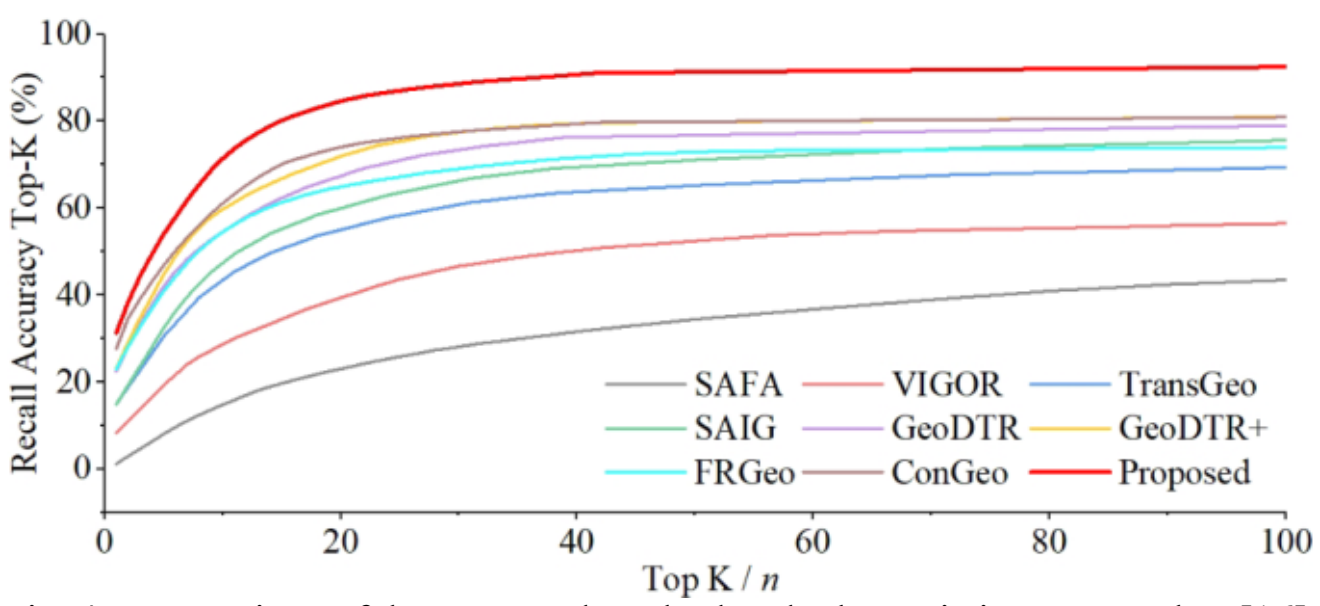

Fig. 4. Comparison of the proposed method and other existing approaches [16], [24]–[30]: All models are trained on KITTI dataset.

TABLE IV
ABLATION STUDY OF THE ROLE OF OUR METHOD IN KITTI AND NCP-INTERSECTION. (BOLD: BEST)

| Dataset | Module A | Module B | Module C | r@1 | r@5 | r@10 | r@1% |
|---|---|---|---|---|---|---|---|
| KITTI | √ | × | × | - | - | - | 2.51 |
| | √ | √ | × | 1.57 | 3.25 | 5.92 | 13.74 |
| | √ | × | √ | - | 2.58 | 4.63 | 7.19 |
| | √ | √ | √ | **31.35** | **52.92** | **72.44** | **91.53** |
| NCP-Intersection | √ | × | × | - | - | - | 1.62 |
| | √ | √ | × | - | 2.29 | 4.75 | 10.16 |
| | √ | × | √ | - | 1.83 | 3.60 | 5.41 |
| | √ | √ | √ | **20.65** | **39.81** | **54.73** | **77.09** |

The experimental results presented in Tables I and II demonstrate that with the enhancement to a 64-channel LiDAR, the proposed algorithm achieves optimal performance, outperforming all benchmarked algorithms in the comparative analysis. As the density of point clouds diminishes, GeoISF exhibits a marginal decline in performance metrics, while it consistently surpasses the majority of the parallel algorithms. In comparison, the inherent limitations of ground images in adapting to varying lighting and weather conditions render them less robust compared to ground point clouds. In GeoISF, the modal discrepancies between point clouds and satellite images are effectively alleviated through the construction of instance semantic trees. As evidenced by the performance on the KITTI dataset presented in Table I, the proposed algorithm surpasses all other parallel algorithms in performance at a LiDAR density of 64 channels.

In addition, LiDAR offers a distinct advantage by enabling panoramic scanning and extraction of structural features surrounding the robot in contrast to forward-facing cameras. Consequently, we performed ablation studies on the LiDAR's field of view (FoV), evaluating the performance at FoV settings of 90 degrees, 180 degrees, and 360 degrees using the KITTI dataset, as detailed in Table III. The proposed algorithm exhibits its poorest performance at a FoV of 90 °, achieving merely 31.08% for r@1% and 2.29% for r@1. This suboptimal performance is attributed to the restricted FoV, which significantly constrains the range of road structure extraction and semantic segmentation. Consequently, the spatial representation capability around the robot is substantially diminished, further impeding the effective construction of a semantic forest. With the expansion of the FoV, the accuracy of the algorithm exhibits a consistent improvement. Specifically, for the r@1% metric, the performance at FoV=360°is 1.34 times that at FoV=180°and 2.94 times that at FoV=90°. These findings underscore the sensitivity of the proposed algorithm to the FoV range, and they confirm that optimal results are achieved when the FoV is set to 360°.

The qualitative performance of our algorithm on the two datasets is illustrated in Fig. 2 and Fig. 3. The experimental dataset encompasses a large-scale satellite image spanning approximately 14km × 7km. The proposed approach establishes similarity by efficiently retrieving the correspondence between ground point clouds and localized satellite imagery, thereby enabling the precise localization of the geo-localization within a large-scale satellite image. In addition, we conducted experiments to evaluate recall accuracy in top-k scenarios on the KITTI dataset, as illustrated in Fig. 4. The results demonstrate that the proposed algorithm converges faster and outperforms other algorithms, achieving an impressive r@1% of 91.53%.

### C. *Ablation Study*

The proposed method consists of three modules: 1) ground semantic segmentation; 2) construction of an instance semantic forest; and 3) feature matching. Therefore, we also conduct ablation experiments on these three modules to verify the benefits of their combinations. In Table IV, modules A, B, and C represent feature matching, construction of an instance semantic forest, and ground semantic segmentation. Based on experiments conducted on two datasets, it is evident that the absence of an instance semantic forest in either ground or satellite images has a significant impact on localization accuracy. The optimal localization effect can only be achieved when all three modules are operated simultaneously. Therefore, although the functions of the three modules differ, they are all essential for ensuring the accuracy and robustness of the localization effect.

## V. Conclusion

This paper presents a novel approach to large-scale ground-to-satellite geo-localization based on point clouds, addressing the critical challenges of semantic alignment and modality gaps in cross-view localization. The proposed method introduces an innovative instance semantic forest constructed using WordNet, which enhances temporal semantic representation and discriminative power by integrating semantic trees from multiple frames. By leveraging environmental semantic representation as a shared medium, our approach effectively bridges the modality gap between point clouds and satellite images, significantly improving semantic matching accuracy. Extensive experimental results demonstrate the superior performance of our algorithm in large-scale cross-view localization, offering a robust solution to the accuracy challenges inherent in such scenarios.

Currently, a significant shortage exists in cross-view localization methods that utilize LiDAR and satellite images. This scarcity is primarily attributed to the challenges associated with feature alignment stemming from cross-modal input. This paper attempts to narrow this gap by leveraging the semantic forest from a ground-to-air perspective in largescale scenarios. The proposed method not only advances the state-of-the-art in this domain but also opens up new avenues for future research, particularly in the areas of computational efficiency and adaptive localization in dynamic environments. In our future endeavors, we strive to advance 3-DoF cross-view pose estimation tasks by incorporating additional ground semantics in large-scale outdoor environments.